\documentclass[lettersize,journal]{IEEEtran}
\IEEEoverridecommandlockouts

\usepackage{cite}
\usepackage{amsmath,amssymb,amsfonts}
\usepackage{algorithmic}
\usepackage{graphicx}
\usepackage{textcomp}
\usepackage[dvipsnames, table]{xcolor}
\usepackage{paralist}
\usepackage[inline]{enumitem}
\usepackage{caption}
\usepackage{makecell}
\usepackage{tabu, booktabs}
\usepackage{float}
\usepackage{subcaption}
\usepackage{comment}
\usepackage{todonotes}
\usepackage[breaklinks,colorlinks,bookmarks=true]{hyperref}
\usepackage{csquotes}
\usepackage{soul}
\usepackage{pifont}
\usepackage{multirow}
\usepackage{tcolorbox}
\usepackage{tabularx}
\usepackage{array}
\usepackage{colortbl}
\usepackage{siunitx}
\usepackage{svg}

\definecolor{magma_darker}{HTML}{fdc38a}
\definecolor{magma_dark}{HTML}{e15666}
\definecolor{magma_light}{HTML}{82247f}
\definecolor{magma_lighter}{HTML}{1f0c43}

\definecolor[named]{xBlue}{HTML}{18647E}
\definecolor[named]{xOrange}{HTML}{FF9B00}
\definecolor[named]{xGray}{HTML}{808080}
\definecolor[named]{xGreen}{HTML}{60B950}
\definecolor[named]{xRed}{HTML}{A30B37}
\definecolor[named]{xDarkBlue}{cmyk}{1,0.58,0,0.21}

\definecolor{cadetblue}{rgb}{0.37, 0.62, 0.63} % A variant of cadet blue
\definecolor{grayish}{rgb}{0.6, 0.6, 0.6} % A variant of grey
\definecolor{redorange}{rgb}{0.91, 0.41, 0.17} % A variant of red orange
\definecolor{limegreen}{rgb}{0.5, 0.8, 0.2} % A variant of lime green

\hypersetup{
    colorlinks=true,
    linkcolor=xOrange,
    filecolor=magenta,      
    urlcolor=xDarkBlue,
    citecolor=xBlue,
}

\usepackage[a4paper, total={184mm,239mm}]{geometry}
\def\BibTeX{{\rm B\kern-.05em{\sc i\kern-.025em b}\kern-.08em
    T\kern-.1667em\lower.7ex\hbox{E}\kern-.125emX}}
    
\ExplSyntaxOn
\DeclareExpandableDocumentCommand{\convertlen}{ O{cm} m }
 {
  \dim_to_decimal_in_unit:nn { #2 } { 1 #1 } cm
 }
\ExplSyntaxOff
\usepackage{gnuplottex}
\usepackage{tikz}
\usepackage{mathtools}

\usetikzlibrary{matrix,calc,positioning,arrows.meta}

\newcommand*\circled[1]{\tikz[baseline=(char.base)]{\node[shape=circle,fill,inner sep=0.5pt] (char) {\textcolor{white}{#1}};}}
\newcommand*\pilar[3]{\tikz[baseline=(char.base)]{\node[shape=circle,fill=#2,minimum size=2.1ex,inner sep=0pt] (char) {\textcolor{#3}{#1}};}}

\begin{document}

\title{Ecomap: Sustainability-Driven Optimization of Multi-Tenant DNN Execution on Edge Servers}

\author{
\IEEEauthorblockN{Varatheepan~Paramanayakam$^{1}$, Andreas Karatzas$^{1}$, Dimitrios Stamoulis$^2$, Iraklis Anagnostopoulos$^1$}~\\
\IEEEauthorblockA{$^1$School of Electrical, Computer and Biomedical Engineering, Southern Illinois University, Carbondale, IL, U.S.A.}
\IEEEauthorblockA{$^2$Department of Electrical and Computer Engineering, The University of Texas at Austin, Austin, TX, U.S.A.}
\IEEEauthorblockA{Email: \{varatheepan, andreas.karatzas, iraklis.anagno\}@siu.edu, dstamoulis@utexas.edu}
}

% \author{
% \IEEEauthorblockN{Andreas Karatzas}
% \IEEEauthorblockA{\textit{Electrical, Computer and Biomedical Engineering} \\
% \textit{Southern Illinois University}\\
% Carbondale, U.S.A. \\
% andreas.karatzas@siu.edu}
% \and
% \IEEEauthorblockN{Iraklis Anagnostopoulos}
% \IEEEauthorblockA{\textit{Electrical, Computer and Biomedical Engineering} \\
% \textit{Southern Illinois University}\\
% Carbondale, U.S.A. \\
% iraklis.anagno@siu.edu}
% }

\maketitle
\begin{abstract}

Edge computing systems struggle to efficiently manage multiple concurrent deep neural network (DNN) workloads while meeting strict latency requirements, minimizing power consumption, and maintaining environmental sustainability. This paper introduces Ecomap, a sustainability-driven framework that dynamically adjusts the maximum power threshold of edge devices based on real-time carbon intensity. Ecomap incorporates the innovative use of mixed-quality models, allowing it to dynamically replace computationally heavy DNNs with lighter alternatives when latency constraints are violated, ensuring service responsiveness with minimal accuracy loss. Additionally, it employs a transformer-based estimator to guide efficient workload mappings. Experimental results using NVIDIA Jetson AGX Xavier demonstrate that Ecomap reduces carbon emissions by an average of 30\% and achieves a 25\% lower carbon delay product (CDP) compared to state-of-the-art methods, while maintaining comparable or better latency and power efficiency. 
\end{abstract}


\setlist{nosep}

\section{Introduction}

The rapid adoption of Artificial Intelligence (AI) and Deep Neural Network (DNN)-based applications has revolutionized numerous fields, including healthcare, autonomous systems, and smart cities~\cite{kwon2021heterogeneous}. These applications require substantial computational resources to deliver real-time responses and meet strict latency requirements. However, this rapid growth in computation has raised environmental concerns due to the carbon emissions produced while running these systems~\cite{gupta2021chasing}. 

The energy consumed during the operation of AI systems generates carbon emissions, known as operational emissions. The carbon footprint (CF) of these operations can be calculated by multiplying the energy consumed by the Carbon Intensity (CI) of the energy source~\cite{lindberg2021guide,gupta2021chasing}. Carbon intensity measures how much carbon dioxide (CO$_2$) is emitted per unit of electricity consumed, which varies according to the energy mix of the power grid, the time of day, and the geographical location. For example, coal-generated electricity has a much higher carbon intensity than renewable sources such as wind or solar. This makes operational emissions a crucial metric for assessing AI systems' environmental impact, especially in edge computing scenarios. Prior works emphasized the importance of carbon awareness throughout the lifecycle of ML and the importance of finding an optimal balance between performance and carbon emissions in the deployment stage~\cite{panteleaki2024carbon}. In particular, edge computing faces unique challenges, as it must balance energy-efficient operation with demands for fast response times and continuous availability. Since edge servers typically rely on local power grids with varying carbon intensities, optimizing their energy usage improves both performance and environmental impact.

Edge computing is becoming an essential paradigm for modern applications, with its adoption growing exponentially due to its ability to process data closer to end users, thereby reducing latency and bandwidth requirements~\cite{kim2023greenscale}. Unlike cloud computing, edge systems primarily handle real-time inference tasks with strict timing requirements. While cloud systems can reduce carbon emissions by efficiently grouping jobs through temporal and spatial batching~\cite{ahmad2021challenges}, these techniques would introduce unacceptable delays in edge computing. This fundamental difference necessitates new sustainability strategies tailored to edge computing's unique energy and performance demands.

Furthermore, modern edge servers are increasingly required to execute machine learning services concurrently, creating resource conflicts that increase latency and reduce responsiveness~\cite{dagli2024shared}. Given the limited computational resources of edge servers, simple coarse-grain methods that map entire DNNs onto a single processing unit~\cite{kang2020scheduling} are highly suboptimal. Instead, effective edge computing requires sophisticated runtime managers that can split DNN layers and allocate resources synergistically across all available computing components, such as CPUs and GPUs, to maximize system performance~\cite{karatzas2023omniboost}. The challenges are further augmented by the vast design space, which requires advanced search algorithms to identify optimal solutions efficiently~\cite{karatzas2024rankmap,karatzas2023omniboost}.

Additionally, optimizing for operational carbon emissions adds a new layer of complexity to an already challenging problem. Runtime managers must carefully control the edge server's energy consumption since energy use is directly linked to carbon emissions. However, merely reducing energy consumption is insufficient. Prior studies have demonstrated that focusing exclusively on first-order metrics, such as power or energy consumption, does not always lead to reduced operational carbon emissions~\cite{gupta2022act}. The complexity is further amplified when considering strategies like lowering operating frequencies or deactivating computational components (e.g., CPU cores) to save energy. Such adjustments alter the dynamics of the device, often invalidating previously optimal mappings of workloads across computational resources. For instance, reducing the frequency of a GPU to save power may increase contention and unbalance resource utilization, leading to degraded performance and higher latency. This cascading effect necessitates continuous re-optimization of the workload mapping, making the management of multi-DNN workloads even more intricate. Therefore, \emph{there is a pressing need for advanced edge-based runtime managers that can synergistically utilize computing components, dynamically adjust to changes in power configurations, minimize latency, and reduce operational carbon emissions simultaneously.}

In scenarios where multiple machine learning services run simultaneously, resource contention can significantly increase latency, adversely impacting user experience~\cite{karatzas2024balancing}. A promising solution to address this challenge is the concept of \emph{mixed-quality} models, which involves using variations of the same AI model architecture with differing sizes, computational demands, and accuracy levels~\cite{stamoulis2019single}. This approach enables runtime managers to dynamically adjust the computational workload by selecting the appropriate model variant based on current resource availability and system constraints.
Consider a video surveillance system, where ResNet-50 serves as the backbone for object detection. During high contention scenarios, such as when other services run concurrently, the runtime manager can replace ResNet-50 with a lighter variant like ResNet-38. This switch significantly reduces computational load and latency while maintaining acceptable detection accuracy. Studies have shown that such transitions typically result in small and acceptable accuracy loss~\cite{li2023clover}, making this strategy highly practical. Mixed-quality models can serve as a valuable control knob for sustainability oriented runtime management in two critical scenarios:
\begin{inparaenum}[\bgroup\bfseries(1)\egroup]
\item when resource contention arises from concurrent execution of DNNs, and
\item when operating frequencies are reduced to save energy and minimize operational carbon emissions.
\end{inparaenum}
However, there is currently no systematic approach that integrates synergistic fine-grain mapping, dynamic power management, and the adaptation of mixed-quality models to effectively reduce operational carbon emissions in edge servers.

In this paper, we present \textbf{Ecomap}, a sustainability-oriented framework for managing multi-DNN workloads on heterogeneous edge servers while meeting strict latency requirements. The main goal of Ecomap is to balance performance and sustainability in edge environments, where low latency is crucial for ensuring a good user experience. To achieve this, Ecomap employs a transformer-based multi-DNN mapping manager that performs power-aware, fine-grained layer-splitting. Additionally, it leverages the concept of \textit{mixed-quality models} to dynamically adapt workloads, enabling the system to meet latency constraints while optimizing resource utilization and reducing carbon emissions. \textbf{The core contributions of Ecomap are threefold:}
\begin{inparaenum}
    \item[\circled{1}] It employs a fine-grained layer-splitting approach to synergistically map concurrent DNNs across available computing components, such as the CPU and GPU. This reduces resource contention and improves system efficiency, significantly lowering latency.
    \item[\circled{2}] Ecomap dynamically adjusts the device's power consumption by fine-tuning the operational frequencies of the CPU and GPU and by controlling the number of active CPU cores. These adjustments are guided by real-time carbon intensity, enabling the system to reduce operational emissions while maintaining adequate performance.
    \item[\circled{3}] Ecomap dynamically utilizes mixed-quality models to adjust the computational workload of running tasks. By replacing high-quality models with lightweight alternatives under resource contention or power constraints, Ecomap ensures that latency constraints are met without significant degradation in accuracy or user experience.
\end{inparaenum}
\emph{This integration of mixed-quality models, power management, and fine-grained workload mapping makes Ecomap a comprehensive solution for sustainable multi-DNN management in edge systems.}

\section{Related Work}\label{sec:related}

\textbf{Multi-DNN execution on resource constrained devices:}
Efficiently utilizing the heterogeneity of resource-constrained devices has been a focus of several studies. For example, the work in~\cite{hsieh2019case} explores inter-layer parallelism in DNNs to optimize throughput but does not address power or energy efficiency. Similarly, the authors in~\cite{wang2019high} propose a linear correlation between the execution time of CNN layers and the dimensions of the matrices involved to map layers more effectively, but they do not consider power and sustainability. To better utilize the heterogeneous components of edge devices, the authors in~\cite{baek2020multi} developed a latency estimation model for DNN pipelines, aiming to improve system throughput. HaX-CoNN~\cite{dagli2024shared} introduces a shared memory contention-aware scheduling framework for running concurrent DNN workloads on heterogeneous SoCs. However, like earlier works, this model ignores power efficiency and sustainability. ODMDEF~\cite{lim2021odmdef} uses linear regression and $k$-nearest neighbors to create pipelines for multi-DNN workloads, but it requires a large dataset to achieve acceptable accuracy and does not consider power efficiency. Other studies focus on specific optimizations. For instance, the authors in~\cite{liu2021cartad} propose an RL-based framework that employs DVFS on multicore systems for efficient scheduling. However, their work targets thermal optimization rather than the co-optimization of power and throughput. Similarly, ARM-CO-UP~\cite{aghapour2024arm} increases throughput via sub-DNN pipelining for consecutive input frames but does not address the concurrent execution of multiple DNNs. OmniBoost~\cite{karatzas2023omniboost} is one of the first frameworks to use a neural network as a cost model, but it does not consider power consumption in its optimization goals. MapFormer~\cite{karatzas2024mapformer} enables fine-grained layer-splitting to improve system throughput and reduce power consumption. However, its approach is conservative in managing power consumption and incurs significant runtime overhead, limiting its suitability for real-time requests.

\textbf{Mixed-quality ML models:} The concept of mixed-quality ML models has been studied a lot, from traditional DNNS to Large Language Models (LLMs)~\cite{paramanayakam2024less}. The work in~\cite{wang2023mixed} introduces a scheme for progressive bit-width allocation and joint training to optimize mixed-precision quantized networks under multiple compression rates. Similarly, the authors in~\cite{motetti2024joint} propose a unified framework that combines pruning and mixed-precision quantization to improve latency and reduce memory usage in DNNs. AutoMPQ~\cite{xu2024autompq} takes a different approach by employing an automatic mixed-precision neural network search method, using a few-shot quantization adapter to adjust the bit-width of each layer dynamically based on specific requirements. Edge-MPQ~\cite{zhao2024edge}, on the other hand, introduces a hardware-aware, layer-wise mixed-precision quantization strategy aimed at optimizing DNN inference on edge devices, striking a balance between accuracy and efficiency. In contrast, the approach in~\cite{spantidi2022targeting} utilizes a wide range of computing components with different precisions, but its heuristic is not suitable for conventional embedded devices. In the context of sustainability, Clover~\cite{li2023clover} presents a runtime system designed to reduce carbon emissions in large-scale ML inference services. By leveraging mixed-quality models and GPU resource partitioning, Clover balances performance, accuracy, and emissions, although its focus is limited to cloud infrastructure. Similarly, PULSE~\cite{sankaranarayanan2024pulse} employs mixed-quality models to optimize the cost of maintaining serverless functions in a ``keep-alive'' state. It dynamically switches between high- and low-quality model variants based on workload demand, effectively balancing latency and resource efficiency while reducing operational overhead.

\textbf{Sustainability-oriented edge computing:} Several works have focused on carbon-aware strategies to enhance sustainability in computing systems. For cloud-based environments, a carbon-aware scheduler is proposed in~\cite{hanafy2024going}, which balances carbon emissions, performance, and cost to achieve significant carbon savings with minimal performance overhead. Similarly, the benefits of scheduling workloads during periods of low-carbon energy availability are explored in~\cite{wiesner2021let}, where a publicly available simulation framework evaluates the potential of carbon-aware scheduling algorithms across different regions. However, these approaches are primarily designed for cloud infrastructures and do not address the challenges of edge computing. GreenScale~\cite{kim2023greenscale} introduces a carbon-aware framework for optimizing edge-cloud infrastructures by modeling carbon emissions based on workload characteristics, renewable energy availability, and runtime variability. This enables efficient scheduling to reduce the carbon footprint of edge applications. In the context of IoT environments, the authors in~\cite{yang2024carbon} propose a carbon-aware dynamic task offloading (CADTO) algorithm for NOMA-enabled mobile edge computing systems. Similarly, LSCEA-AIoT~\cite{song2024carbon} is a low-carbon sustainable computing framework designed to optimize energy-efficient data acquisition and task offloading in AIoT ecosystems. For DNN workloads, CarbonCP~\cite{ke2024carboncp} employs conformal prediction theory for context-adaptive, carbon-aware DNN partitioning, focusing on edge-cloud offloading scenarios. However, these frameworks do not address the specific challenges of optimizing multi-DNN workloads on heterogeneous edge servers, which require advanced methods for synergistic resource allocation and carbon-aware runtime management.
\section{Background}\label{sec:background}

In this section, we provide background on operational emissions, carbon footprint, and the correlation with carbon intensity, along with their temporal and spatial characteristics.

\subsection{Operational Emissions}

The environmental impact of edge computing systems is measured through operational emissions - the carbon footprint generated during system operation. These emissions depend on two key factors: the energy consumed and its carbon intensity, which varies by source. Each energy source produces a distinct amount of CO$_2$ per kilowatt-hour (kWh) of electricity generated. Non-renewable sources like coal (820 gCO$_2$/kWh), oil (650 gCO$_2$/kWh), and natural gas (490 gCO$_2$/kWh) have significantly higher emission rates than renewable alternatives. In contrast, renewable sources like wind (11 gCO$_2$/kWh), nuclear (12 gCO$_2$/kWh), and hydro (24 gCO$_2$/kWh) produce far fewer emissions, making them crucial for sustainable edge computing deployments~\cite{gupta2022act}.

\subsection{Carbon Intensity and Regional/Time Variability} 

Carbon intensity ($CI$) represents the average carbon dioxide emissions per unit of electricity generated, serving as a critical metric for assessing the environmental impact of energy consumption.
Mathematically, carbon intensity is expressed as:
\begin{equation}\label{eq:ci}
CI = \frac{\sum_{i=1}^N E_i \cdot CEF_i}{\sum_{i=1}^N E_i}
\end{equation}
where $E_i$ is the electricity generated by source $i$ (measured in kWh), $CEF_i$ represents the carbon emission factor of source $i$ (in gCO$_2$/kWh), and $N$ is the total number of electricity generation sources in the region. Equation~\ref{eq:ci} highlights that the carbon intensity of electricity depends on both the quantity of energy generated per source and its respective emission factor.

\begin{figure}
    % \centering
    \begin{subfigure}{\columnwidth}
        \includegraphics[width=0.99\columnwidth]{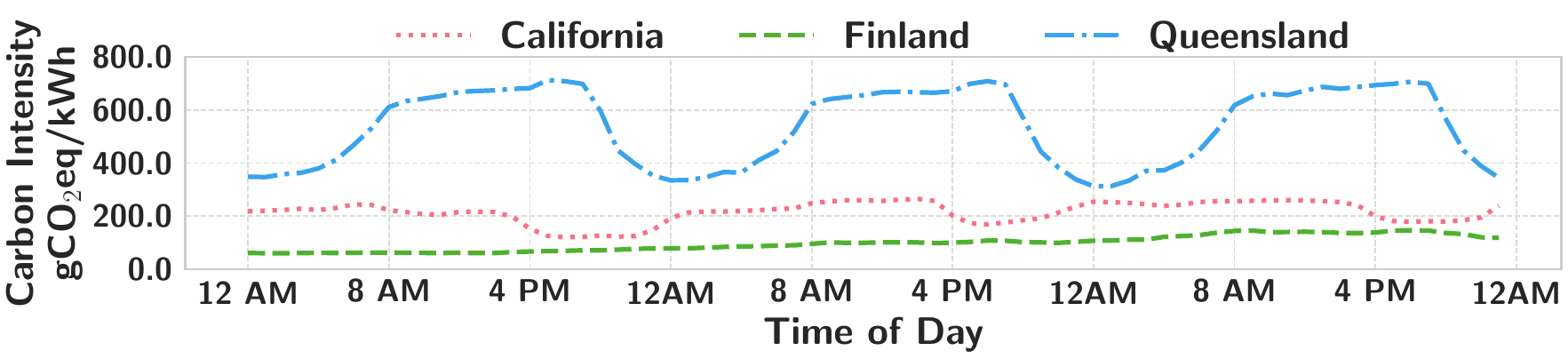}
        \captionsetup{justification=centering}
        \caption{Regional carbon intensity variation}
        \label{fig:regional_ci}
    \end{subfigure}
    \begin{subfigure}{\columnwidth}
        \includegraphics[width=0.99\columnwidth]{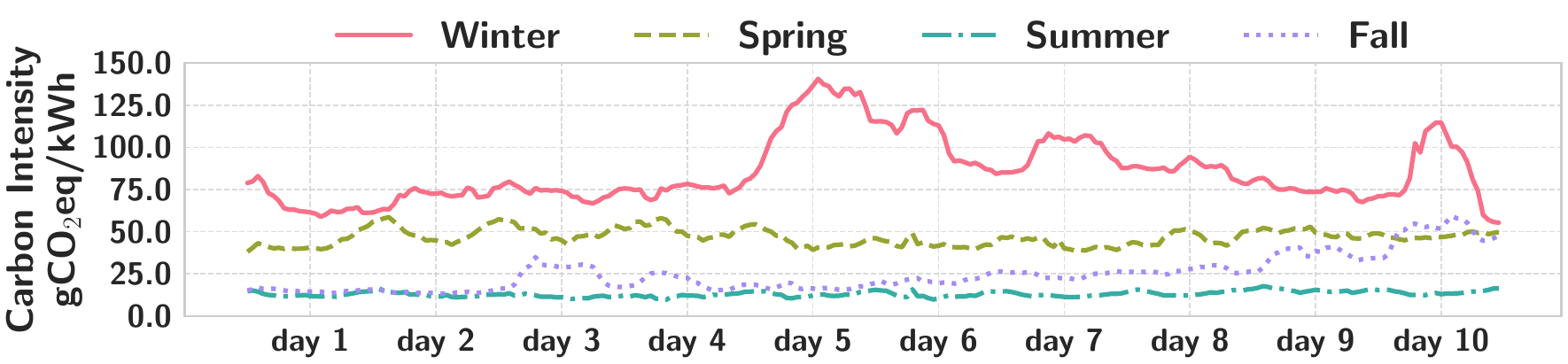}
        \captionsetup{justification=centering}
        \caption{Seasonal carbon intensity variation in Finland}
        \label{fig:seasonal_ci}
    \end{subfigure}
    \begin{subfigure}{\columnwidth}
        \includegraphics[width=0.99\columnwidth]{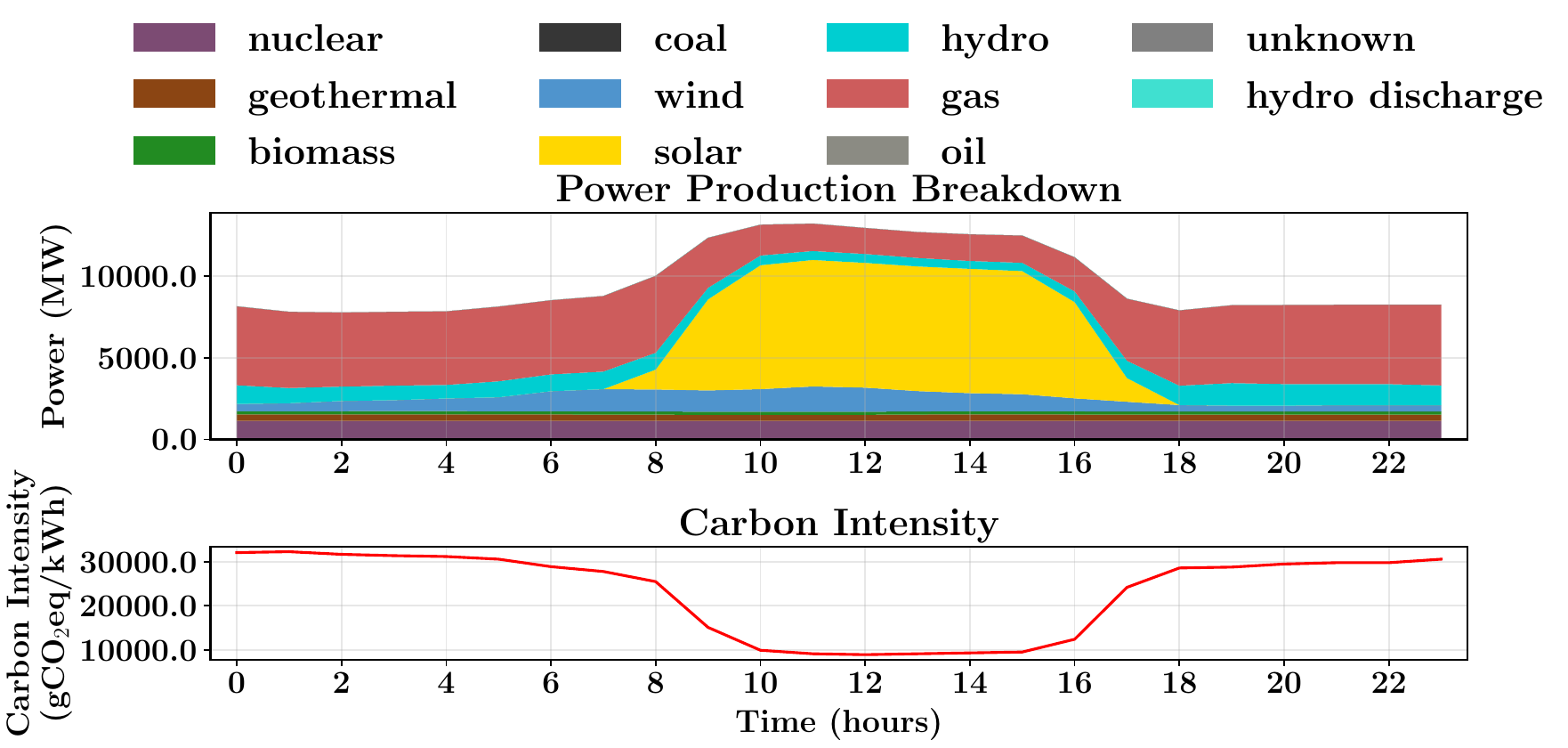}
        \captionsetup{justification=centering}
        \caption{CI variation with power source composition}
        \label{fig:composition_ci}
    \end{subfigure}
    \caption{Examples of variation of Carbon Intensity over geographical locations, time, and energy mix of the power grid. Data taken from~\cite{electricitymaps}.}    
    \label{fig:CI}
\end{figure}

The carbon intensity of a region's electricity grid depends on the mix of energy sources and their availability. Regions with a high proportion of renewable energy, such as wind or solar power, tend to have lower carbon intensity. Conversely, regions heavily dependent on coal or natural gas, exhibit much higher carbon intensity due to the high emission factors of these non-renewable sources. Carbon intensity also varies over time, driven by fluctuations in energy demand and the availability of renewable energy. Solar power, for instance, peaks during daylight hours, significantly reducing carbon intensity in regions with substantial solar capacity. However, during periods of high energy demand, such as evenings or cold winters, non-renewable sources like natural gas often supplement renewable energy to meet the load, leading to a temporary increase in carbon intensity. 
Figure~\ref{fig:CI} shows an example of how $CI$ varies over different geographical areas (Figure~\ref{fig:regional_ci}), seasons (Figure~\ref{fig:seasonal_ci}), and mix of energy sources (Figure~\ref{fig:composition_ci}).

\subsection{Operational Emissions and Carbon Intensity}

The operational emissions, expressed as $CF$, of a system are directly proportional to its energy consumption and the carbon intensity of the electricity it uses~\cite{gupta2022act}: 
\begin{equation}
    CF = E \times CI
\end{equation}
where $CF$ represents the operational emissions of the system, measured in grams of CO$_2$ (gCO$_2$), $E$ denotes the energy consumed by the system, expressed in kilowatt-hours (kWh), and $CI$ is the carbon intensity of the electricity source, quantified in grams of CO$_2$ per kilowatt-hour (gCO$_2$/kWh). Minimizing $CF$ in edge computing systems presents unique challenges compared to centralized cloud environments, primarily due to the real-time inference tasks that edge computing supports.
\section{Methodology}\label{sec:methodology}

\begin{figure*}
    \centering
    \resizebox{1\textwidth}{!}{\includegraphics[page=5, width=\linewidth, clip, trim={2em 4 3 3}]{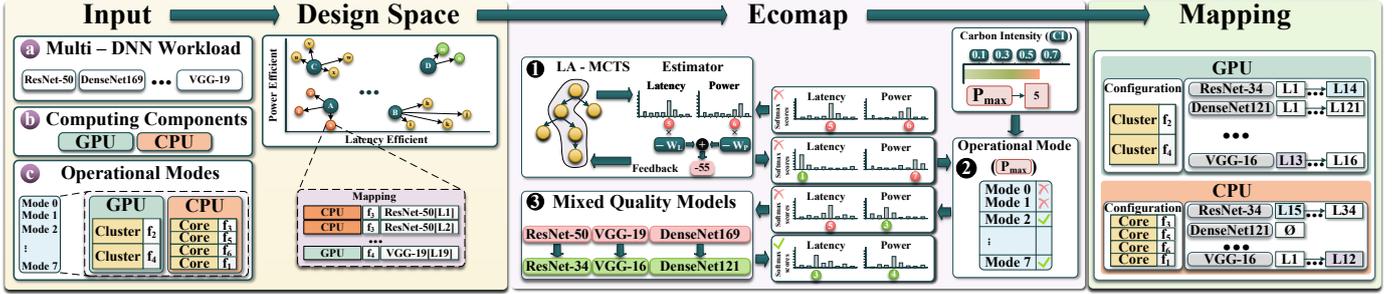}}
    \caption{Overview of our proposed framework: Our key insight behind Ecomap is to dynamically define maximum power threshold for the edge server based on \emph{current} carbon intensity of the grid, ensuring operation emission reduction while meeting performance requirements.}
    \label{fig:overview}
\end{figure*}

Ecomap is a sustainability-driven framework designed to optimize the execution of multiple DNN workloads on heterogeneous edge servers while meeting strict latency constraints. \emph{A key feature of Ecomap is its ability to define a dynamic maximum power threshold ($P_{\text{max}}$) for the edge server based on the current carbon intensity ($CI$) of the electricity grid}. This dynamic adaptation ensures that the system reduces operational emissions while meeting performance requirements. Figure~\ref{fig:overview} provides a high-level overview of the Ecomap framework.

\textbf{\ul{Input and design space:}} Ecomap takes as inputs: \pilar{\textbf{a}}{Plum}{white} A set of DNNs to be executed concurrently; \pilar{\textbf{b}}{Plum}{white} The available computing components on the edge server; and \pilar{\textbf{c}}{Plum}{white} A list of hardware operational modes. Each mode represents a specific hardware configuration, defined by parameters including the number of active CPU cores, CPU/GPU frequencies, and memory frequency, with an associated maximum power threshold (Section~\ref{sec:part1-operating_modes}). These inputs create a vast design space of possible mappings and configurations. We employ the Latent Action Monte Carlo Tree Search (LA-MCTS) algorithm to efficiently explore this space within a computational budget. LA-MCTS leverages a transformer-based estimator to accurately predict throughput and power consumption for each candidate mapping, enabling accurate ranking of solutions (Sections~\ref{sec:estimator}-\ref{seq:la-mcts}).

\textbf{\ul{Runtime:}} At runtime, Ecomap dynamically determines the maximum operational power threshold ($P_{\text{max}}$) for the edge server based on real-time carbon intensity ($CI$) of the electricity grid. This dynamic power threshold ensures that the system adapts to changing environmental conditions, reducing operational emissions while keeping up with performance requirements (Section~\ref{sec:runtime}). Once $P_{\text{max}}$ is calculated, Ecomap uses its transformer-based estimator and LA-MCTS to identify an optimal mapping of DNN workloads to available hardware resources and select an operational mode. This process aims to minimize service delay while ensuring that the power consumption remains within the threshold dictated by the given $CI$, achieving a balance between sustainability and performance.

\textbf{\ul{Enabling mixed-quality models:}} Ecomap also continuously monitors the latency and power performance of the running services. If latency thresholds are violated, the framework leverages the concept of mixed-quality models. It replaces computationally intensive DNNs with lighter variants from the same family to reduce delays without significant loss in accuracy (Section~\ref{sec:mixed-aulity}). This adaptive strategy ensures that service-level agreements (SLAs) are met even under dynamic workloads and environmental conditions (changes in $CI$).

\subsection{Operating modes}\label{sec:part1-operating_modes}

We define device-specific hardware operational modes to control power consumption. Each operational mode corresponds to a specific hardware configuration, defined by parameters such as the number of active CPU cores and the frequencies of the CPU, GPU, and memory. These modes are precomputed and stored in a lookup table (LUT), which is used at runtime (Section~\ref{sec:runtime}) to control the maximum power consumption of the device and thereby reduce operational emissions. 

\begin{table}[H]
\centering
\caption{Operating modes}
% \vspace*{-10pt}
\label{tab:operating_modes}
\resizebox{0.8\columnwidth}{!}{
\begin{tabular}{|l|c|c|c|c|c|}
\hline
\textbf{$m_i$} & $\mathbf{c}$ & $\mathbf{f_{\text{CPU}}}$ & $\mathbf{f_{\text{GPU}}}$ & $\mathbf{f_{\text{mem}}}$ & $\mathbf{P_{\text{max}}}$ \\ \hline
1 & 8 & 2.2GH & 1.3GH & 2.1GH & 30W \\
2 & 6 & 2.2GH & 1.3GH & 2.1GH & 26W \\
3 & 4 & 2.2GH & 1.3GH & 2.1GH & 22W \\
4 & 8 & 1.8GH & 828MH & 2.1GH & 16W \\
5 & 6 & 1.8GH & 828MH & 2.1GH & 13W \\
6 & 4 & 1.8GH & 828MH & 2.1GH & 11W \\
7 & 8 & 1.2GH & 675MH & 1.2GH & 8W \\
8 & 6 & 1.2GH & 675MH & 1.2GH & 6W \\
\hline
\end{tabular}}
\end{table}

We denote the LUT of operational modes as $\mathcal{M} = {m_1, \dots, m_k}$, where $m_i$ is an operational mode, and $k$ is the total number of modes. Each $m_i$ can be described by a tuple:
\begin{equation}
    m_i = (c, f_{\text{CPU}}, f_{\text{GPU}}, f_{\text{mem}}, P_{\text{max}})
\end{equation}
where $c$ is the number of active CPU cores, $f_{\text{CPU}}$ is the frequency of the CPU cores, $f_{\text{GPU}}$ denotes the frequency of the GPU, $f_{\text{mem}}$ is the memory frequency, and $P_{\text{max}}$ is the achieved maximum power consumption of that mode. These configurations allow Ecomap to adjust the device's power consumption in a fine-grained manner while executing concurrent DNN workloads. For example, on the NVIDIA Jetson AGX Xavier board, we created a lookup table (LUT) consisting of eight operational modes, enabling power consumption to range from 8 W to 30 W in small steps, as detailed in Table~\ref{tab:operating_modes}.

The importance of having precomputed operational modes becomes evident when considering the challenges of dynamically adjusting CPU, GPU, and memory frequencies at runtime to meet specific power thresholds. Suppose that multiple DNNs run concurrently on an NVIDIA Jetson AGX Xavier board, and the system needs to reduce its power consumption from 30 W to 20 W. Without precomputed operational modes, the frequency controller would need to iteratively adjust the frequencies of the CPU and GPU to find a configuration that satisfies the 20 W power cap. This process often involves trial and error, as reducing the frequency of one component (e.g., the GPU) might not be sufficient and could require complementary adjustments to the CPU frequency. Moreover, any changes to the frequencies must consider the utilization levels of these components by the running DNNs and the characteristics of the DNNs themselves. For example, if a latency-critical DNN heavily utilizes the GPU, reducing its frequency could increase latency and violate service-level agreements (SLAs). To compensate, the system might need to remap some GPU workloads to the CPU, but this shift can increase CPU utilization, potentially requiring further adjustments to the CPU frequency. These cascading effects make it extremely challenging to achieve a stable configuration that balances power, latency, and performance.

Thus, by supporting operational modes with well-defined hardware configurations and power caps, Ecomap eliminates this complexity. For instance, when the runtime manager (Section~\ref{sec:runtime}) is transitioning to an operational mode with $P_{\text{max}} = 20$ W, the LUT already provides an optimized configuration that accounts for the expected utilization of the CPU and GPU based on the DNN workloads. This ensures that the system can quickly switch to a mode that satisfies the power constraint while maintaining the performance and latency requirements of the running services.

\subsection{Latency and power estimator}\label{sec:estimator}

As mentioned before, Ecomap takes as input: 
\begin{inparaenum}[(\bgroup\bfseries i\egroup)]
\item a set of DNNs to be executed simultaneously;
\item the set of available computing components; and
\item the supporting operational modes.
\end{inparaenum}
To process this data, we transform it into numerical vector representations using a learnable composite embedding module~\cite{mikolov2013distributed} that incorporates the latent representations of:
\begin{inparaenum}[(\bgroup\bfseries i\egroup)]
\item the computational profile of each DNN layer within the workload,
\item the processing capabilities of each computing component of the embedded device, and
\item the number and operational frequency of all computing components for each mode $m_i$.
\end{inparaenum}
Ecomap utilizes layer partitioning to break down any DNN model into smaller sub-DNNs, requiring a layer-level input representation. To that end, for each layer in the workload, we apply our tailored embedding module to create a sequence of tuples, each consisting of a layer, a computing component, and its corresponding operational frequency in mode $m_i$. Unlike previous methods~\cite{karatzas2023omniboost,lim2021odmdef}, our distributed embedding vectors are learnable, enhancing the transformer's ability to estimate latency and power consumption more accurately. Transformers do not inherently understand tokens' relative or absolute positions in a sequence, so we incorporate a standard sinusoidal positional encoding layer~\cite{takase2019positional}. 

Building on the structure of our input sequence $\mathcal{S}$, we use a casual transformer-based estimator~\cite{vaswani2017attention} to assess any mapping $\mathcal{M}$ and predict its latency and power consumption under each different mode $m_i$. The choice of a transformer-based estimator is due to its ability to identify long-sequence numerical patterns, which is crucial for managing higher-order multi-DNN workloads—specifically, workloads where DNNs have more than 1,000 fine-grained partitions to be mapped. Estimators from previous studies~\cite{karatzas2023omniboost,lim2021odmdef,kang2020scheduling}, although effective for smaller workloads, tend to underperform with larger multi-DNN workloads, often resulting in sub-optimal mappings. 

A major differentiator of Ecomap from previous state-of-the-art approaches is that it is designed for a classification rather than a regression task. Specifically, our transformer-based estimator predicts quantile distributions of latency and power consumption scores. While estimating exact values for these metrics could potentially yield better multi-DNN mappings, it also requires significantly larger datasets to manage the imbalances in target values~\cite{thabtah2020data}. For instance, mappings that achieve low latency scores are relatively rare compared to those with higher latency, creating an imbalance in the dataset that can lead to inaccurate predictions. To address this, we define the estimator's target as a distribution of $N$ discrete classes, i.e., quantiles, effectively transforming the problem into a classification task. The $N$ quantiles are equal in sample size, which helps overcome data imbalance. Furthermore, to manage the multi-objective nature~\cite{hu2021unit} of predicting both latency and power consumption, we feed the contextualized sequence outputs from the transformer encoder into two separate fully connected layers, each with $N$ neurons corresponding to the number of classes in our target distribution.

\subsection{LA-MCTS module}\label{seq:la-mcts}

Our estimator module is the mechanism for evaluating any candidate mapping. Therefore, we still need a design space exploration mechanism. To address the exploration of the mappings, we integrate the Latent Action-MCTS (LA-MCTS)~\cite{wang2020learning} algorithm, a highly efficient space exploration module. MCTS is a heuristic approach that efficiently navigates extensive design spaces by iteratively interacting with its decision tree within a set computational budget~\cite{swiechowski2023monte}. This tree holds all possible mappings for a given design space. Although traditional MCTS effectively minimizes a cost function through stochastic processes, it tends to converge slowly. This slow convergence increases both the computational workload and the number of required estimator inferences.

\begin{figure}[H]
    \centering
    \resizebox{1\columnwidth}{!}{\includegraphics[page=6, width=\linewidth, clip, trim={2em 4 3 3}]{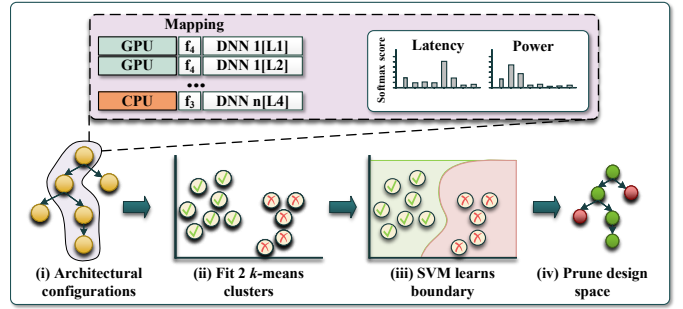}}
    \caption{Design space pruning via LA-MCTS.}
    \label{fig:method:space-pruning}
\end{figure}

To enhance the convergence rate of MCTS, we adopted LA-MCTS, which iteratively learns to partition the design space hierarchically. In each iteration, LA-MCTS examines specific regions of the decision tree and applies a $k$-means algorithm to categorize them into two clusters, distinguishing between promising (good) and less promising (bad) solutions. It uses Support Vector Machines (SVM)~\cite{patle2013svm} to create a decision boundary that extrapolates the patterns identified by the $k$-means to the broader design space. This process helps prioritize the most promising regions of the design space by assigning a likelihood score to each candidate mapping, indicating its potential for further consideration. Figure~\ref{fig:method:space-pruning} provides a high-level overview of the iterative process of LA-MCTS and how it prunes the design space to focus on more viable solutions.

To address the multi-objective nature of our problem, we formulate a composite value function $\mathcal{V}$. This function evaluates any mapping $\mathcal{M}$ by calculating the weighted difference between the predicted latency class and the predicted power consumption class. Additionally, to ensure the satisfaction of the power constraint ($\mathcal{P}^{threshold}$), we incorporate a filtering step based on the predicted power consumption class. Specifically, if a mapping is estimated to exceed the maximum allowable power consumption, its value is set to negative infinity, effectively removing it from consideration as a viable solution. This approach is detailed mathematically in Equation~\ref{eq:method:reward-calc}.
\begin{equation}
    \mathbf{V}(\mathcal{M}) = \begin{cases} 
    \mathbf{W}_{L} \cdot \mathcal{L}(\mathcal{M}) - w_2 \cdot \mathbf{W}_{P} \cdot \mathcal{P}(\mathcal{M}), & \\ 
    \qquad \text{ if } \mathcal{P}(\mathcal{M}) \leq \mathcal{P}^{threshold} \\
    -\infty, \text{otherwise}
    \end{cases}
    \label{eq:method:reward-calc}
\end{equation}
Here, $\mathbf{W}_{L}$ represents the weight assigned to latency, $\mathcal{L}(\mathcal{M})$ denotes the estimated latency class for the mapping $\mathcal{M}$, $\mathbf{W}_{P}$ is the weight assigned to power consumption, and $\mathcal{P}(\mathcal{M})$ indicates the predicted power consumption. The maximum allowable power consumption, determined by $CI$ (Section~\ref{sec:part1-operating_modes}) is denoted by $\mathcal{P}^{threshold}$.

\subsection{Runtime}\label{sec:runtime}

At runtime, Ecomap incorporates a 24-hour $CI$ prediction for the electricity grid to dynamically manage the device's operational power thresholds. Using the predictive method presented in~\cite{maji2022carboncast}, which achieves high accuracy for daily $CI$ forecasting, Ecomap ensures that its decisions are proactive. Based on this forecast, Ecomap determines the minimum ($CI_{\text{min}}^{\text{day}}$) and maximum ($CI_{\text{max}}^{\text{day}}$) values of $CI$ over the next 24 hours. The period associated with $CI_{\text{min}}^{\text{day}}$ represents the optimal time for high-power operations, as operational emissions have the lowest environmental impact during this window.

When $CI$ is at its minimum, the edge device operates at the highest power threshold, corresponding to the maximum operational mode ($m_1$ in Table~\ref{tab:operating_modes}). This configuration allows the system to deliver services with minimal delays and maximum performance while taking advantage of the low environmental impact during this period. As $CI$ increases throughout the day, Ecomap dynamically adjusts the maximum allowable power threshold ($P_{\text{max}}$) by transitioning to operational modes with finer-grained power settings. The decision to use granular power thresholds, as shown in Table~\ref{tab:operating_modes}, is crucial for maintaining a balance between sustainability and performance. Large gaps between power thresholds could result in abrupt changes that might either overcommit resources, causing unnecessary emissions, or undercommit resources, leading to latency violations. By defining operational modes with small, incremental differences in $P_{\text{max}}$, Ecomap ensures smooth transitions between configurations, enabling more precise control of power consumption while adapting to varying carbon intensity levels. These fine-grained adjustments allow the system to remain responsive to changes in $CI$ without significant disruptions to service performance. To avoid erratic behavior (e.g., frequent back-and-forth changes in $P_{\text{max}}$), Ecomap updates the power threshold only when $CI$ changes by at least 10\% of the predicted range. This threshold-based mechanism ensures stable and efficient runtime operation, mapping the $CI$ ranges to the corresponding operating modes.

Additionally, Ecomap accounts for the arrival of new services at any time. For each new service, Ecomap uses its latency and power estimator along with the LA-MCTS module to identify a mapping that satisfies the power threshold determined by the current $CI$. Since multiple valid mappings can meet the power constraints, Ecomap employs the reward function described in Subsection~\ref{seq:la-mcts} to prioritize configurations with the lowest latency. This approach balances the need for high performance while satisfying the dynamic power and carbon intensity constraints.

\subsection{Enabling mixed-quality models}\label{sec:mixed-aulity}

Ecomap ensures that all running services meet predefined latency and power thresholds by actively monitoring their performance in real time. Changes in the maximum allowable power threshold ($P_{\text{max}}$), driven by variations in carbon intensity ($CI$) or the arrival of new service requests, can lead to resource contention and latency violations. To address these issues, Ecomap dynamically adapts by leveraging mixed-quality models, which replace computationally intensive DNNs with lighter alternatives from the same model family varying in size, number of layers, or parameter complexity. These alternatives, referred to as mixed-quality models, offer reduced computational requirements while maintaining acceptable accuracy. This adaptability enables Ecomap to sustain service quality under constrained power budgets.

Ecomap monitors services continuously and detects latency violations triggered by changes in $P_{\text{max}}$ or the addition of new service requests. When a latency violation is detected, Ecomap executes the following structured process. First, Ecomap begins by identifying the service experiencing the highest latency relative to its threshold. Let $S$ denote this service, with its associated DNN represented as $D$. Ecomap replaces $D$ with the next available lightweight alternative from the set of mixed-quality models, $M(D) = {D^1, D^2, \dots, D^m}$. The selection is guided by the following optimization:
\begin{equation}
\text{Find } D_i^k \in M(D_i) \text{ such that } L(D_i^k) \leq L_{\text{max}} \text{ and } \Delta A(D_i^k) \leq \epsilon,
\end{equation}
where $L(D^k)$ is the latency of $D^k$, $L_{\text{max}}$ is the maximum allowable latency, $\Delta A(D^k)$ is the accuracy drop of $D^k$ compared to $D$, and $\epsilon$ is the maximum acceptable accuracy drop. If latency constraints are still not met after the first replacement, Ecomap iterates through the remaining alternatives in $M(D)$ until either the violation is resolved or all alternatives are exhausted. If the latency violation persists after exhausting all alternatives for the impacted service, Ecomap identifies the most computationally intensive service in the workload. The DNN for this service is then replaced with a lightweight alternative to reduce contention and free up resources for other services.

A key enhancement in Ecomap is its use of tailored search to ensure that these adaptations occur efficiently. Instead of conducting a full LA-MCTS exploration to find the new mapping, which would involve searching the entire configuration space, Ecomap narrows the search to configurations directly affected by the updated DNN. During training, Ecomap identifies patterns of behavior for each DNN by evaluating performance under various mapping configurations. Regions where layer splitting leads to latency increases exceeding 30\% are deprioritized or excluded, forming a refined search space.

At runtime, this tailored search significantly reduces computational overhead, as it focuses on high-probability configurations while avoiding suboptimal areas. This strategy is particularly effective because in this scenario Ecomap adjusts \emph{only one DNN at a time}, ensuring that the tailored search remains fast and precise. The latency and power estimator evaluates potential mappings within this refined space to select a configuration that satisfies all constraints.

By dynamically adapting services through mixed-quality models and leveraging tailored search, Ecomap maintains latency compliance even under dynamic workloads and environmental conditions. This process allows Ecomap to balance latency, power efficiency, and sustainability, providing an efficient solution for managing multi-DNN workloads in edge computing environments.

\begin{table}
\centering
\caption{Supported services and mixed-quality models}
\label{tab:services_models}
\resizebox{1.0\columnwidth}{!}{
\begin{tabular}{|l|l|l|}
\hline
\textbf{Service}                      & \textbf{Default DNN (Level-1)} & \textbf{Mixed-Quality Models}           \\ \hline
Object Detection                 & MNASNet1\_3                   & MNASNet1\_0, MNASNet0\_75                                  \\ \hline
Object Classification            & EfficientNet\_v2\_s           & EfficientNet\_b1, EfficientNet\_b3                         \\ \hline
Object Tracking                  & ResNet152                     & ResNet101, ResNet50                                        \\ \hline
Depth Estimation                 & ResNet152                     & ResNet101, ResNet50                                        \\ \hline
Abnormal Behavior Detection     & VGG19                         & VGG16, VGG13                                               \\ \hline
Facial Expression Recognition    & DenseNet169                   & DenseNet161, DenseNet121                                   \\ \hline
\end{tabular}}
\end{table}

\section{Experimental Evaluation}\label{sec:experimental}

In this section, we evaluate the performance of Ecomap across key metrics, including latency, power consumption, operational emissions, and sustainability efficiency. This evaluation is performed using the Nvidia Jetson AGX Xavier (JAX) edge server, a state-of-the-art platform that employs
\begin{inparaenum}[(\bgroup\bfseries i\egroup)]
\item a Volta GPU with 512 CUDA cores and 64 Tensor cores rendering a performance peak of 10 TFLOPS;
\item a Carmel CPU with $\times 4$ ARMv8.2 dual-core clusters operating at $2.26 GHz$; and
\item a $32 GB$ LPDDR4x memory.
\end{inparaenum}These hardware capabilities make it an ideal testbed for exploring Ecomap’s effectiveness under various workloads and $CI$ conditions.

Ecomap is developed using PyTorch, which supports the integration of diverse DNN architectures and enables fine-grained partitioning of multi-DNN workloads. To manage these workloads, we created a custom PyTorch-powered compute library that enables dynamic mapping of DNNs onto the edge server's computing components. For the training phase of Ecomap's estimator, we generated a dataset comprising $8,000$ different mappings. To boost the accuracy of the estimator, we included $1,000$ samples for each hardware operational mode. Each workload consists of random combinations of $5$ to $10$ DNNs, executed across the pre-defined operational modes of the device. To ensure a comprehensive evaluation, we leveraged a large set of models available in the \texttt{torchvision.models} library, resulting in a total space of $50$ widely used DNNs. These models are categorized into the following families:
\begin{inparaenum}[(\bgroup\bfseries i\egroup)]
\item AlexNet,
% \item ConvNeXt,
\item DenseNet,
\item EfficientNet,
\item GoogLeNet,
\item InceptionV3,
\item MNASNet,
% \item MaxVit,
\item MobileNetV2,
\item MobileNetV3,
\item RegNet,
\item ResNet,
\item ShuffleNetV2,
\item SqueezeNet,
% \item SwinTransformer,
\item VGG, and
% \item VisionTransformer. 
\end{inparaenum}
Ecomap's design ensures compatibility with most of the models defined in PyTorch, making it adaptable to diverse application requirements. We trained our estimator for $100$ epochs with $80 \%$ of our dataset using \texttt{AdamW} optimizer with $0.0001$ learning rate and \texttt{CosineAnnealingLR} scheduler for smooth approximation of the most optimal model parameter set. For validation, we evaluated our estimator on the remaining and unseen test subset.

For our experiments, we utilized three distinct 5-day periods to evaluate the server's performance under varying carbon intensity ($CI$) conditions and workloads. Week-1 and Week-3 exhibit significant variability in $CI$, reflecting fluctuating energy grid dynamics, whereas Week-2 demonstrates relatively stable $CI$ with minimal fluctuation. These scenarios allow us to test Ecomap's adaptability to different environmental and operational conditions.

Regarding user-based service requests, we tested six types of services:
\begin{inparaenum}[(\bgroup\bfseries i\egroup)]
    \item object detection,
    \item object classification,
    \item object tracking,
    \item depth estimation,
    \item abnormal behavior detection, and
    \item facial expression recognition.
\end{inparaenum}
Each service supports mixed-quality models to ensure adaptability under latency violations. Table~\ref{tab:services_models} shows the default DNN (level-1) used for each service and the mixed-quality models (level-2 and level-3) employed when latency thresholds are exceeded.

Each week also varies in the number of service requests received by the server. In Weeks 1 and 2, the maximum number of requests the server could handle without significant delays or becoming unresponsive was capped at 15 concurrent service instances. For Week-3, the maximum number of requests was reduced to 10 to evaluate system performance under medium-to-heavy workloads. The weekly characteristics, including $CI$ variability and workload intensity, are summarized in Table~\ref{tab:weekly_characteristics}.

\begin{table}[h]
\centering
\caption{Weekly experiment characteristics}
\label{tab:weekly_characteristics}
\begin{tabular}{|c|c|c|}
\hline
\textbf{Week Name} & $\mathbf{CI}$ \textbf{Variability} & \textbf{Workload Intensity} \\ \hline
Week-1            & High                        & High                        \\ \hline
Week-2            & Low                         & High                        \\ \hline
Week-3            & High                        & Medium                      \\ \hline
\end{tabular}
\end{table}

In our experiments, we evaluated Ecomap under two latency thresholds for each service running on the edge server: a relaxed threshold of 2 seconds and a strict deadline of 500 milliseconds. These thresholds reflect different quality-of-service requirements, allowing us to assess Ecomap's ability to balance latency and sustainability under varying constraints. To differentiate between these configurations, we refer to Ecomap operating under the \ul{r}elaxed constraint as $Ecomap_R$ and under the \ul{s}trict constraint as $Ecomap_S$ in the following analysis.

\begin{figure}[h]
    \centering
    \resizebox{1\columnwidth}{!}{\includegraphics[width=\linewidth, clip]{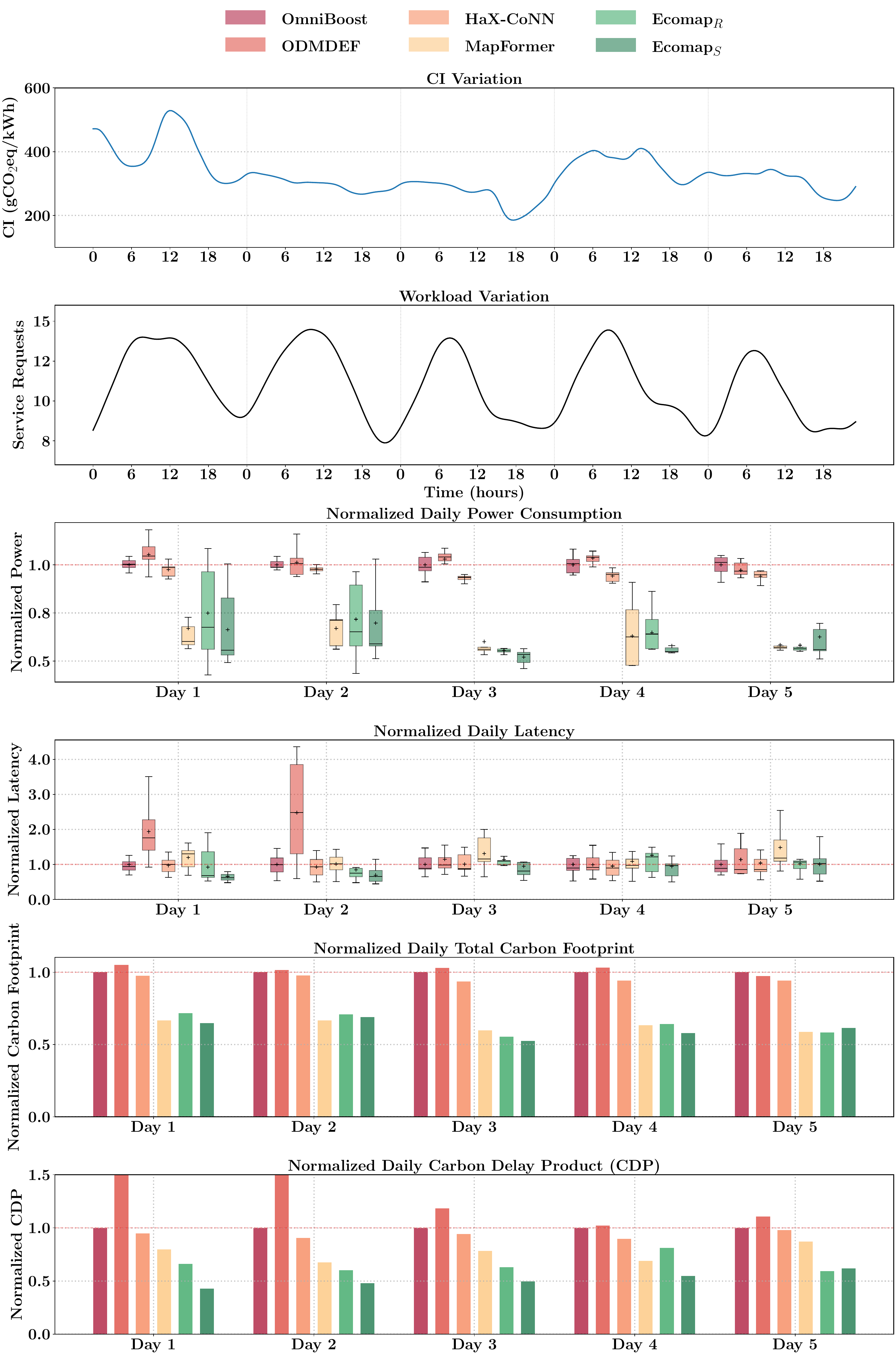}}
    \caption{Normalized comparative analysis of Ecomap ($Ecomap_R$ and $Ecomap_S$) during Week-1, evaluating $CI$ variability, workload distribution, power consumption, latency, daily emissions, and Carbon Delay Product (CDP) over a 5-day period. For all comparison charts, lower is better.}
    \label{fig:res-week1}
\end{figure}

To provide a comprehensive evaluation, we compared Ecomap against several state-of-the-art frameworks for managing multi-DNN workloads on edge servers: \begin{inparaenum}[(\bgroup\bfseries i\egroup)] \item \textbf{OmniBoost}\cite{karatzas2023omniboost}, a greedy throughput optimization framework for multi-DNN workloads, which serves as the baseline for comparison; \item \textbf{ODMDEF}\cite{lim2021odmdef}, a manager utilizing a combination of linear regression and $k$-NN classifiers for DNN scheduling; \item \textbf{Hax-Conn}\cite{dagli2024shared}, a contention-aware scheduling framework designed for concurrent DNN execution; and \item \textbf{MapFormer}\cite{karatzas2024mapformer}, a power-efficient framework aimed at optimizing resource usage for multi-DNN workloads. \end{inparaenum}

To evaluate the performance of Ecomap, we compared it against all the aforementioned methods using a comprehensive set of metrics. Specifically, we measured latency, power consumption, daily carbon footprint, and the Carbon-Delay Product (CDP). The daily carbon footprint quantifies the total operational emissions over a 24-hour period, providing a measure of the environmental impact. Finally, the CDP, a product of latency and carbon footprint, offers an integrated metric to evaluate the trade-off between performance and sustainability.

\subsection{Sustainability-oriented comparison}
\begin{figure}[h]
    \centering
    \resizebox{1\columnwidth}{!}{\includegraphics[width=\linewidth, clip]{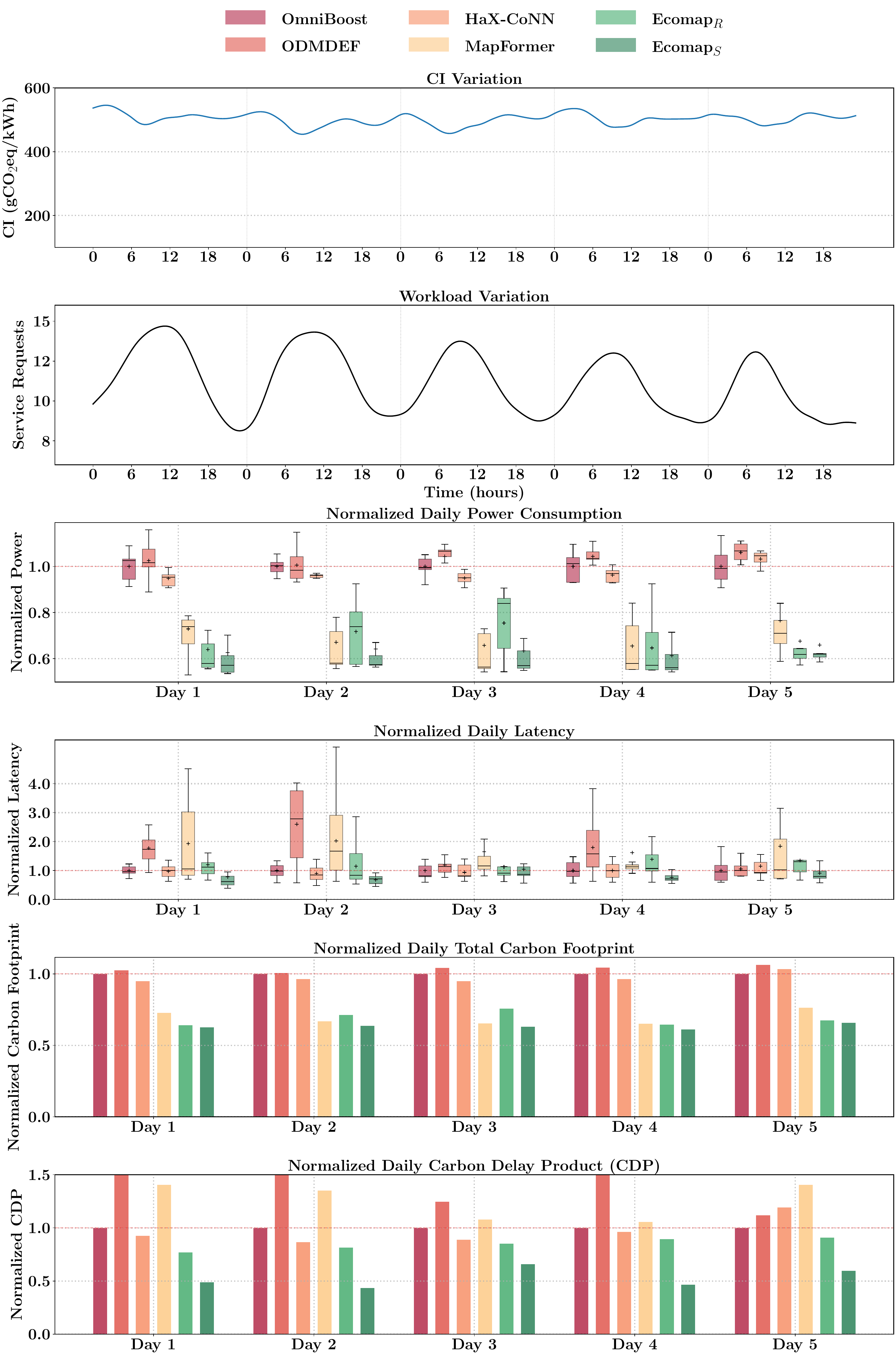}}
    \caption{Normalized comparative analysis of Ecomap during Week-2. For all comparison charts, lower is better.}
    \label{fig:res-week2}
\end{figure}

\begin{figure}[]
    \centering
    \resizebox{1\columnwidth}{!}{\includegraphics[width=\linewidth, clip]{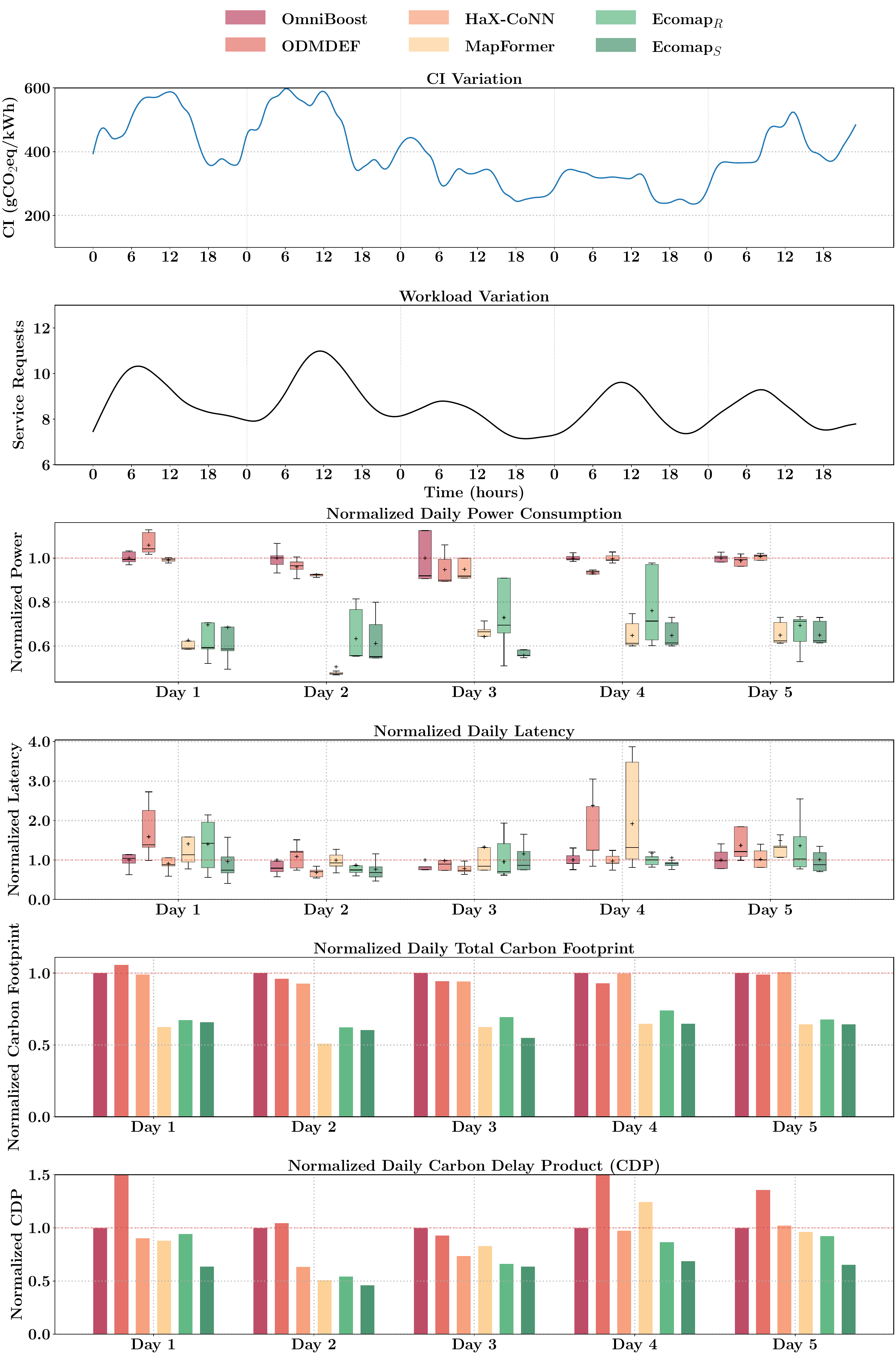}}
    \caption{Normalized comparative analysis of Ecomap during Week-3. For all comparison charts, lower is better.}
    \label{fig:res-week3}
\end{figure}

Figures~\ref{fig:res-week1}-\ref{fig:res-week3} depict the comparison between all methods. Specifically, for Week-1, depicted in Figure~\ref{fig:res-week1}, the $CI$ exhibits significant variability, ranging from approximately 200 gCO$_2$/kWh to 500 gCO$_2$/kWh.~\\
\textbf{\ul{Power consumption:}} Ecomap demonstrates strong power efficiency across both configurations, $Ecomap_R$ and $Ecomap_S$. On average, $Ecomap_R$ reduces power consumption by 35\% compared to OmniBoost and 32\% compared to Hax-Conn, while $Ecomap_S$ achieves slightly lower power consumption and achieves a reduction of 39\% compared to OmniBoost and 32\% compared to Hax-Conn. MapFormer, as expected, remains competitive in terms of power consumption due to its focus on power optimization.~\\
\textbf{\ul{Latency:}}  
Ecomap effectively balances latency in both $Ecomap_R$ and $Ecomap_S$ configurations. $Ecomap_R$ ensures low power while maintaining acceptable service responsiveness. $Ecomap_S$, under the strict 500 ms latency constraint, achieves lower latency values across all days but incurs slightly higher power usage. Compared to OmniBoost and Hax-Conn, $Ecomap_R$ maintains the latency with a slight increase of about 2\%, while $Ecomap_S$ achieves 17\% lower latency due to Ecomap’s dynamic adaptation and mixed-quality models. In contrast, MapFormer performs poorly in terms of latency for both thresholds. Its power-centric design disregards latency requirements, leading to significant delays in real-time services. This comparison underscores Ecomap’s ability to handle multi-DNN workloads effectively under varying latency constraints.~\\
\textbf{\ul{Daily total emissions:}}  Ecomap achieves significant reductions in normalized daily total emissions for both configurations. $Ecomap_R$, benefiting from its relaxed constraints, reduces emissions by 35\% compared to OmniBoost and 33\% compared to Hax-Conn on average across all days. $Ecomap_S$, despite stricter latency requirements, achieves 39\% and 36\% lower emissions than OmniBoost and Hax-Conn, respectively. These results demonstrate Ecomap's effectiveness in minimizing emissions even under challenging operational constraints.~\\
\textbf{\ul{CDP:}}

Ecomap excels in terms of normalized Carbon Delay Product (CDP), which integrates latency and emissions to measure sustainability efficiency. Both $Ecomap_R$ and $Ecomap_S$ outperform MapFormer significantly. $Ecomap_R$ achieves 13\% lower CDP than MapFormer, while $Ecomap_S$ achieves 36\% lower CDP. Despite MapFormer’s exceptional power efficiency, its inability to adapt to latency constraints results in higher latency, which increases its CDP. In contrast, Ecomap leverages mixed-quality models and dynamic threshold adjustments to effectively balance latency and emissions. The difference between $Ecomap_R$ and $Ecomap_S$ indicates that, despite the stricter latency thresholds, the use of mixed-quality models improves both carbon efficiency and latency, making both configurations more sustainable compared to other methods. 
In summary, Ecomap, in both $Ecomap_R$ and $Ecomap_S$ configurations, outperforms state-of-the-art frameworks in terms of power consumption, latency, emissions, and sustainability efficiency. These results highlight Ecomap’s adaptability and ability to balance performance and sustainability in dynamic edge computing environments.

In Week-2 (Figure~\ref{fig:res-week2}), with low $CI$ variability ranging between 450 gCO$_2$/kWh and 550 gCO$_2$/kWh, Ecomap shows strong performance across all metrics. Compared to MapFormer, the \textbf{power consumption} of $Ecomap_R$ has slight increases over several days, but averages on par with a reduction of about 1\%, and $Ecomap_S$ reduces by 9\%. On the other hand, Ecomap completely outperforms OmniBoost and Hax-Conn by 34\% and 32\%, respectively. Ecomap also achieves consistently low \textbf{latency}, with $Ecomap_S$ showing 18\% lower latency than OmniBoost and 16\% lower than Hax-Conn, while MapFormer suffers from significantly higher delays due to its lack of latency awareness. Regarding \textbf{carbon footprint}, $Ecomap_R$ reduces emissions by 32\% compared to OmniBoost and 30\% compared to Hax-Conn, while $Ecomap_S$ achieves reductions of 37\% and 35\%, respectively. Most notably, in terms of \textbf{CDP}, Ecomap outperforms MapFormer significantly, with $Ecomap_R$ achieving 34\% lower CDP and $Ecomap_S$ achieving 60\% lower CDP. 

In Week-3 (Figure~\ref{fig:res-week3}), with medium workload intensity and high $CI$ variability (ranging from 250 to 600 gCO$_2$/kWh), Ecomap continues to have strong performance across all metrics. \textbf{Power consumption} for $Ecomap_R$ remains highly competitive, averaging 10\% higher than MapFormer, while $Ecomap_S$ averages only 2\% higher, both significantly outperforming OmniBoost and Hax-Conn by 34\% and 32\%, respectively. \textbf{Latency} remains low for Ecomap, with $Ecomap_S$ achieving 6\% lower than OmniBoost and maintains the latency around 2\% higher than Hax-Conn, while $Ecomap_R$ maintains comparable latency under relaxed constraints. Ecomap also achieves substantial reductions in normalized daily \textbf{carbon footprint}, with $Ecomap_R$ showing a 33\% reduction compared to OmniBoost and a 31\% reduction compared to Hax-Conn, while $Ecomap_S$ achieves reductions of 38\% and 36\%, respectively. Notably, Ecomap significantly outperforms MapFormer in terms of \textbf{CDP}, with $Ecomap_R$ achieving a 7\% lower CDP and $Ecomap_S$ achieving a 28\% lower CDP.

\subsection{Mixed-quality models analysis}

\begin{figure}[]
    \centering
    \resizebox{1\columnwidth}{!}{\includegraphics[width=\linewidth, clip]{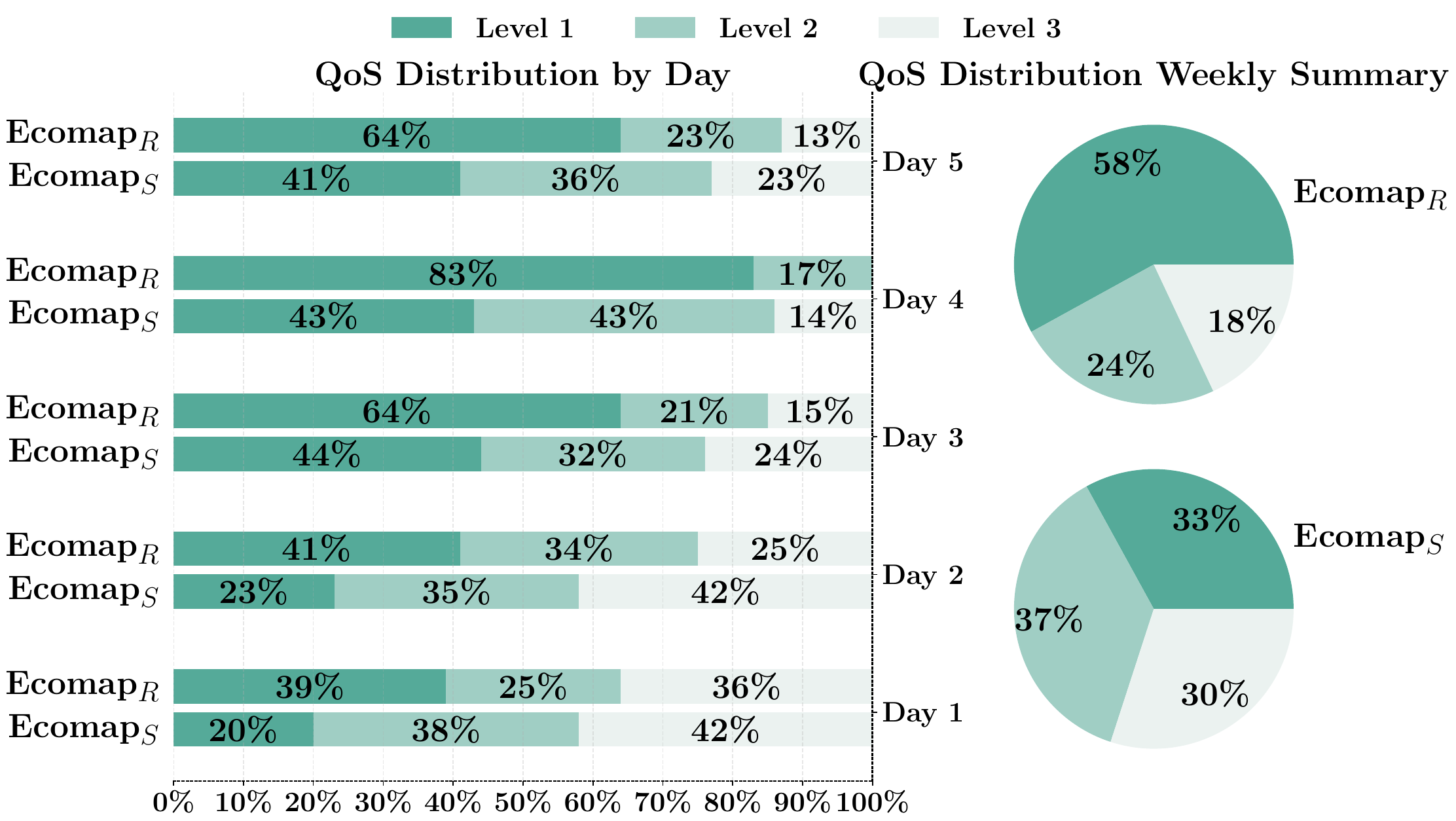}}
    \caption{Distribution of mixed-quality model usage by Ecomap during Week-1.}
    \label{fig:res-week1_qos}
\end{figure}

Mixed-quality models allow Ecomap to adapt to varying latency requirements by replacing high-quality DNNs with lighter alternatives whenever latency constraints are violated. The analysis for Week-1 (Figure~\ref{fig:res-week1_qos}) reveals distinct trends in how $Ecomap_R$ and $Ecomap_S$ utilize mixed-quality models. In general, $Ecomap_R$ relies more heavily on default (level-1) models compared to $Ecomap_S$, which frequently switches to lighter models to meet its stricter constraints. Over the week, $Ecomap_R$ processes an average of 58\% of tasks with default models, while $Ecomap_S$ only achieves 33\%, reflecting the additional adaptations required under tighter latency thresholds. $Ecomap_S$ demonstrates a higher reliance on lightweight models across all days due to the need to meet its stricter threshold, with default models used the least on Day 2, accounting for only 14\% of tasks. 

\begin{figure}[]
    \centering
    \resizebox{1\columnwidth}{!}{\includegraphics[width=\linewidth, clip]{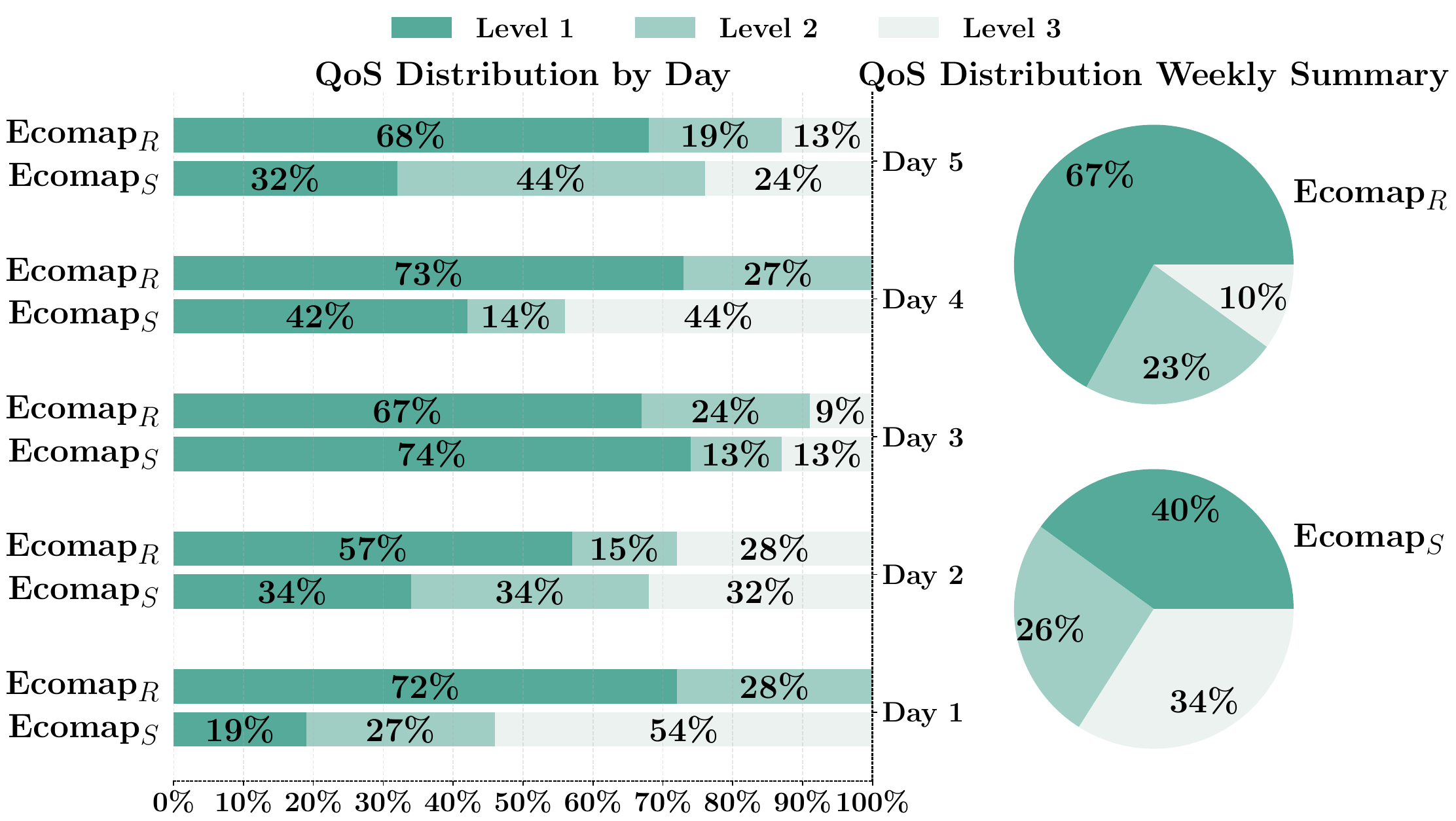}}
    \caption{Distribution of mixed-quality model usage by Ecomap during Week-2.}
    \label{fig:res-week2_qos}
\end{figure}

In Week-2 (Figure~\ref{fig:res-week2_qos}), Ecomap maintains consistent behavior across both configurations, $Ecomap_R$ and $Ecomap_S$, leveraging mixed-quality models effectively to meet latency constraints. For $Ecomap_R$, a significant portion of tasks (averaging $67\%$) is handled by default (level-1) models due to the relaxed latency threshold. In contrast, $Ecomap_S$, under stricter constraints, relies more heavily on lightweight alternatives, with $40\%$ of tasks processed using level-1 models. 

\begin{figure}[]
    \centering
    \resizebox{1\columnwidth}{!}{\includegraphics[width=\linewidth, clip]{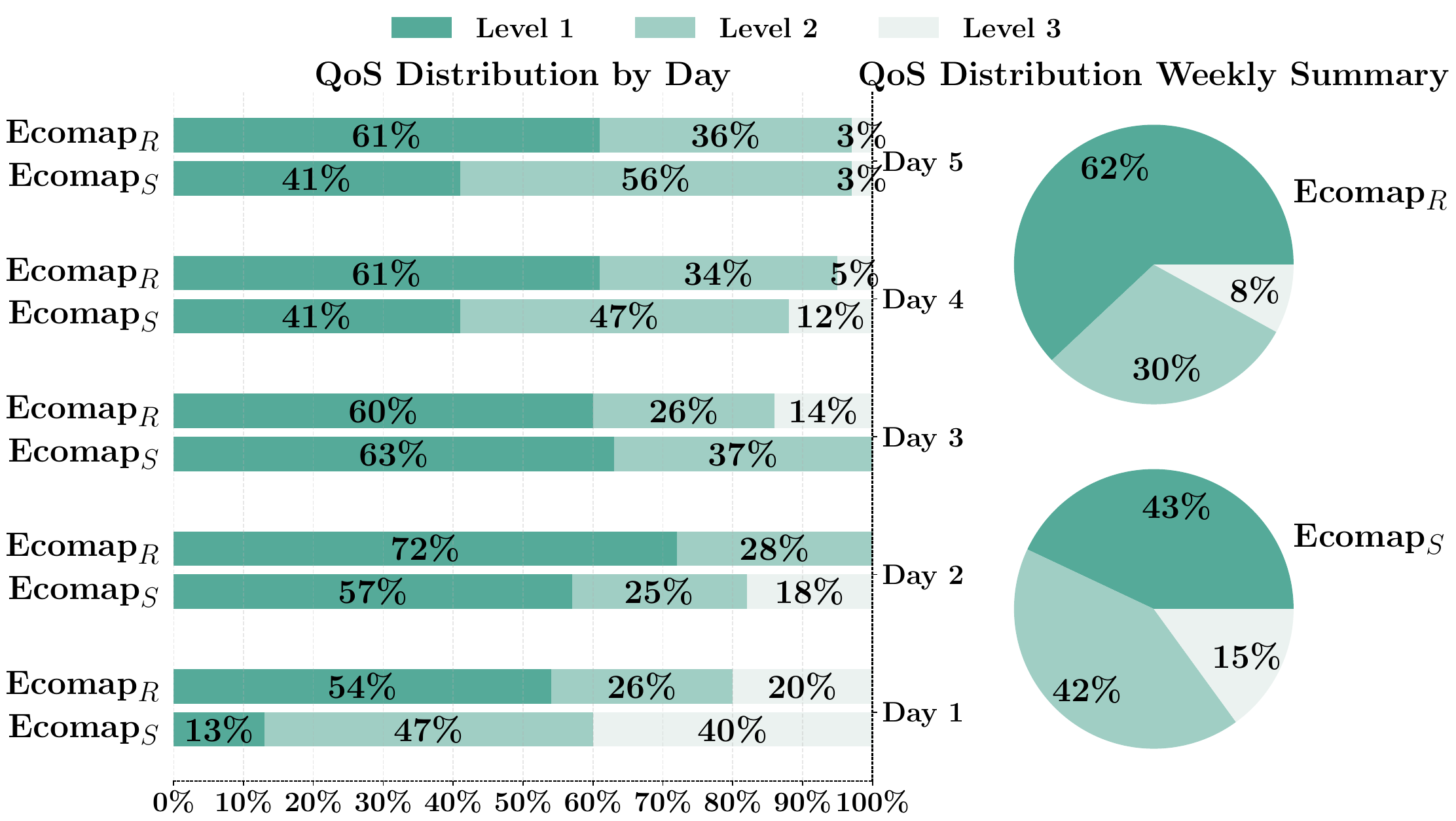}}
    \caption{Distribution of mixed-quality model usage by Ecomap during Week-3.}
    \label{fig:res-week3_qos}
\end{figure}

In Week-3 (Figure~\ref{fig:res-week3_qos}), with a medium workload intensity, there is a noticeable increase in the use of level-1 models for $Ecomap_S$, averaging $43\%$ across the week compared to $33\%$ in Week-1. This shift indicates that the reduced workload intensity allows $Ecomap_S$ to accommodate more tasks with default models while still adhering to its strict latency constraints. The increased use of level-1 models in Week-3 demonstrates how Ecomap efficiently balances workload demands and latency requirements while minimizing the need for lightweight alternatives under less intensive conditions. 

\section{Conclusion}\label{sec:conclusion}

This paper presents Ecomap, a sustainability-driven management framework for multi-DNN workloads on edge devices. Unlike conventional methods that prioritize either throughput or power efficiency, Ecomap dynamically adjusts operational power thresholds based on carbon intensity ($CI$), ensuring a balance between low latency and minimized environmental impact. Our experiments validate Ecomap's better performance in reducing operational emissions and optimizing the carbon delay product across varying workloads and $CI$ conditions. While Ecomap achieves comparable power efficiency to other power-efficient methods, it surpasses them in sustainability by effectively adapting to real-time $CI$ variations, maintaining latency thresholds, and leveraging mixed-quality models for critical scenarios. These findings underline the potential of Ecomap to enable carbon-aware, efficient edge computing.

\section*{Acknowledgments}

This work is supported by grant NSF CCF 2324854.

\bibliographystyle{IEEEtran}
\bibliography{ref}
\begin{IEEEbiography}[{\includegraphics[width=1in,height=1.25in,clip,keepaspectratio]{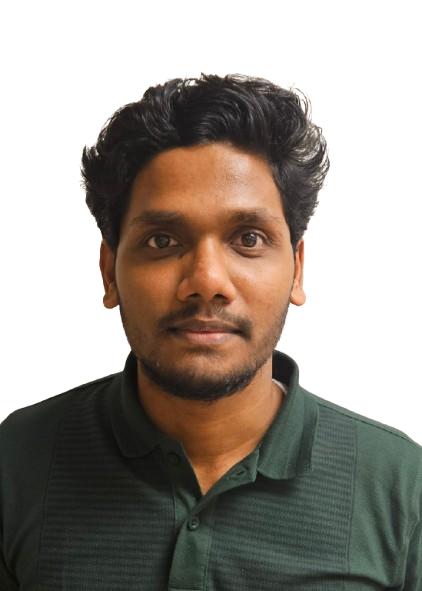}}]
    {Varatheepan Paramanayakam} received the Bachelor of Science of Engineering degree from the Department of Electronic and Telecommunication Engineering, University of Moratuwa, Sri Lanka, in 2021. He is currently pursuing the Doctor of Philosophy degree at the School of Electrical, Computer and Biomedical Engineering at Southern Illinois University, Carbondale, Illinois, as a member of the Embedded Systems Software Laboratory. His research interests include embedded systems, sustainable AI, and deep learning.
\end{IEEEbiography}

\begin{IEEEbiography}[{\includegraphics[width=1in,height=1.25in,clip,keepaspectratio]{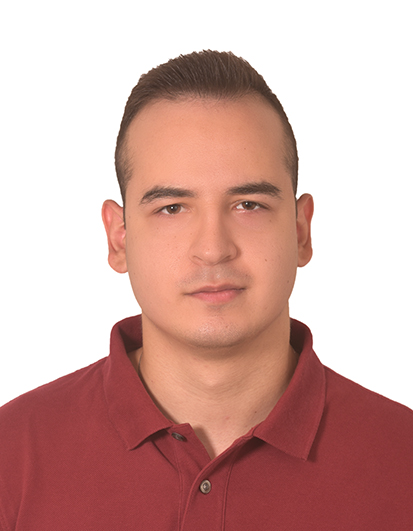}}]
    {Andreas Karatzas} received the Integrated Master degree (Diploma) from the department of Computer Engineering and Informatics (CEID), University of Patras, Patras, Greece, in 2021. He is currently pursuing the Ph.D. degree at the School of Electrical, Computer and Biomedical Engineering at Southern Illinois University, Carbondale, Illinois, as a member of the Embedded Systems Software Lab. His research interests include embedded systems, approximate computing, and deep learning. 
\end{IEEEbiography}

\begin{IEEEbiography}[{\includegraphics[width=1in,height=1.25in,clip,keepaspectratio]{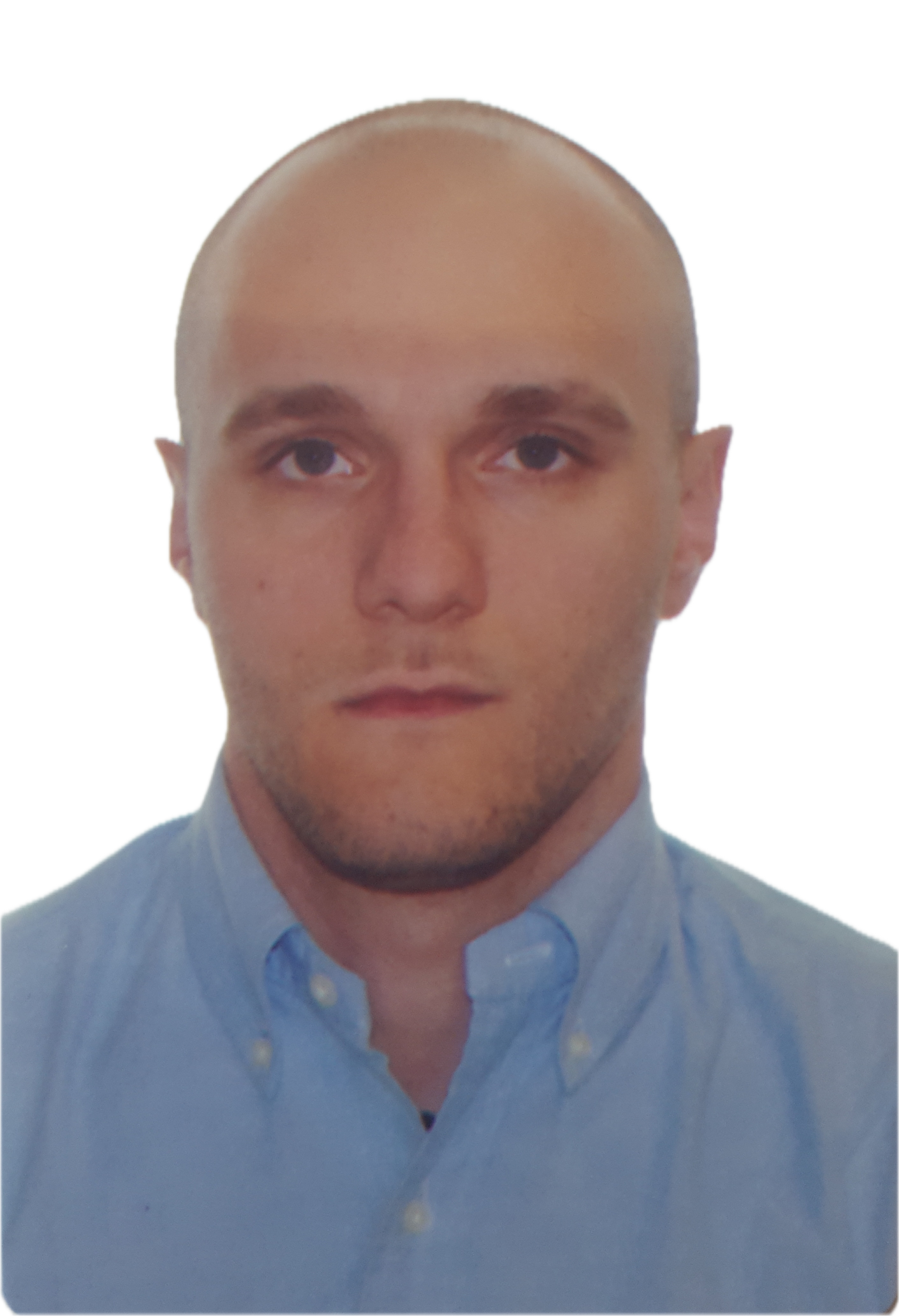}}]
    {Dimitrios Stamoulis} is a Special Faculty member in the Dept. of Electrical and Computer Engineering (ECE) at The University of Texas at Austin, Austin, TX. Previously, he founded and led the \textit{CoStrategist} R\&D Group at Microsoft Mixed Reality. He received his PhD in ECE from Carnegie Mellon University, where he specialized on hardware-aware AutoML. He also holds a MEng in ECE from McGill University and a Diploma in ECE from the National Technical University of Athens.
\end{IEEEbiography}

\begin{IEEEbiography}[{\includegraphics[width=1in,height=1.25in,clip,keepaspectratio]{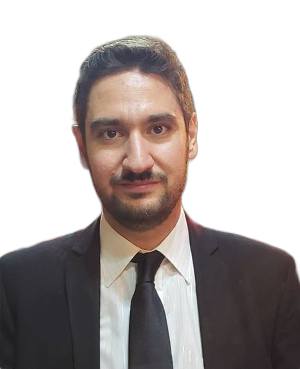}}]
    {Iraklis Anagnostopoulos} is an Associate Professor at the School of Electrical, Computer and Biomedical Engineering at Southern Illinois University, Carbondale. He is the director of the Embedded Systems Software Lab, which works on run-time resource management of modern and heterogeneous embedded many-core architectures. He received his Ph.D. in the Microprocessors and Digital Systems Laboratory of National Technical University of Athens. His research interests lie in the area of machine learning, heterogeneous hardware accelerators, and hardware/software co-design.
\end{IEEEbiography}
\end{document}